\documentclass{article}

\usepackage[accepted]{eiml_icml2026}
\usepackage{amsmath,amssymb,amsfonts}
\usepackage{graphicx}
\usepackage{booktabs}
\usepackage[colorlinks,citecolor=blue,urlcolor=blue,linkcolor=black]{hyperref}

\title{The Posterior-Prior Epistemic Gap: When World Model Imagination Fails to Carry Task-Relevant Signals}

\author{
Jaerock Kwon\textsuperscript{1} \quad
Elahe Delavari\textsuperscript{1} \quad
Donghyun Kim\textsuperscript{2} \quad
Haewoon Nam\textsuperscript{2} \\[4pt]
\textsuperscript{1}Department of Electrical and Computer Engineering,
University of Michigan--Dearborn \\
\textsuperscript{2}Department of Electrical Engineering,
Hanyang University, South Korea \\[2pt]
\texttt{\{jrkwon, elahed\}@umich.edu} \quad
\texttt{\{kissw, hnam\}@hanyang.ac.kr}
}

\date{}

\begin{document}
\maketitle

\begin{abstract}
Model-based planners that roll out imagined futures assume the
transition model faithfully propagates task-relevant signals---a
property rarely measured directly.  In a single-seed case study
of a Recurrent State Space Model world model with active-inference planning in the CARLA driving simulator,
we identify a \textbf{posterior-prior gap}---posterior
(observation-filtered) states retain obstacle information that
imagined rollout states progressively lose: a supervised obstacle
probe achieves AUC 0.85 on posterior states, but the
same probe applied to imagined prior states shows monotonic decay
(AUC 0.85$\rightarrow$0.71 over 15 steps; logit separation drops
57\%).  Because the probe is identical, the decay isolates the
transition model---not the representation---as the information
bottleneck.  Obstacle avoidance drops to 0\% without privileged
runtime coordinates, and the planner's own ensemble-based
epistemic term does not flag the loss---ensemble disagreement yields AUC 0.52 as an obstacle
classifier (chance level), making the failure an unflagged
imagination failure---an unknown unknown at planning
time.\footnote{Project page: \url{https://bimilab.github.io/paper-posterior-prior-epistemic-gap/}}
As a secondary contribution, we show that
\textbf{task-conditioned weighting of Effective Free Energy components} over the
same world model yields lane keeping ($81.2 \pm 1.8$\% route
completion) and obstacle avoidance (28/28 avoided, \emph{using
privileged coordinates}) in the CARLA simulator, with runtime weight modulation
enabling post-evasion centering.
\end{abstract}

%----------------------------------------------------------------------
\section{Introduction}

Model-based planning with learned world
models~\citep{hafner2023mastering} relies on rolling out imagined
futures and scoring candidate action sequences.  Planning quality
therefore depends on whether the transition model faithfully
propagates task-relevant signals through imagination---a property
that is usually assumed but rarely measured directly.

Active inference~\citep{friston2010free, parr2022active} provides
one concrete instantiation: an agent selects actions that minimize
Expected Free Energy (EFE), naturally balancing goal-seeking
(instrumental value) with information-seeking (epistemic value).
We use it here as the planning framework, but the diagnostic we
propose---probing the same auxiliary head on posterior versus
imagined prior states---applies to any world-model-based planner
that distinguishes filtered and predicted latent states.

We study a Recurrent State Space Model (RSSM) world model~\citep{hafner2023mastering} with improved Cross-Entropy Method (iCEM)
planning~\citep{pinneri2021sample} in the CARLA driving
simulator~\citep{dosovitskiy2017carla} and ask: \emph{does the
posterior's competence at encoding task-relevant signals transfer
to the imagined prior that the planner actually uses?}  In this
single-seed case study, our answer is negative.  The filtered
posterior encodes obstacle presence at AUC 0.85, but probing the
same auxiliary head on imagined prior states reveals 57\% logit
separation decay over 15 steps (AUC 0.85$\rightarrow$0.71), and
avoidance remains at 0\%.  Crucially, the planner's own ensemble-based epistemic term
$G_e$ seen in \eqref{eq:efe} does not flag this loss: ensemble disagreement as an
obstacle classifier yields AUC 0.52 (chance level), 
making the failure an unflagged imagination
failure---an unknown unknown at planning time.
Our contributions:

\begin{enumerate}
    \item A \textbf{directly measured posterior-prior gap}
    in RSSM-based planning: probing imagined states shows 57\%
    logit-separation decay over 15 imagination steps
    (AUC 0.85$\rightarrow$0.71), yielding 0\% avoidance without
    privileged information.  The planner's ensemble-based
    epistemic term does not flag this loss (AUC 0.52 as
    obstacle classifier---chance level).

    \item A \textbf{diagnostic probe} that applies the same
    auxiliary obstacle head to both posterior and imagined prior
    states, isolating the transition model---not the
    representation---as the primary bottleneck (with the caveat
    that the probe is trained on posterior states, so
    distribution shift may contribute; see
    Section~\ref{sec:vision}).  This design can be applied
    to other world-model-based planners.

    \item \textbf{Task-conditioned weighting of EFE components}
    over a shared RSSM world model, producing lane keeping
    ($81.2 \pm 1.8$\% route completion) and obstacle avoidance
    (28/28 with \emph{privileged} runtime coordinates) from a
    single objective via runtime weight modulation---demonstrating
    behavioral flexibility, not learned obstacle reasoning.
\end{enumerate}

%----------------------------------------------------------------------
\section{Related Work}

\textbf{World models for control.}
DreamerV3~\citep{hafner2023mastering} uses RSSM imagination for
RL-based policy learning, achieving strong results across diverse
domains. MILE~\citep{hu2022model} learns driving world models for
imitation.  Transformer-based world
models~\citep{chen2022transdreamer, micheli2023transformers}
replace the recurrent transition with attention, potentially
altering how information propagates through multi-step
imagination; whether the posterior-prior gap we identify persists
under attention-based transitions is an open question.
These systems evaluate their world models on reconstruction and
reward prediction, but not on whether safety-critical object
signals survive multi-step imagination rollouts---the angle this
paper takes.

\textbf{Deep active inference.}
\citet{catal2020learning} and \citet{fountas2020deep} apply deep
active inference with VAE world models to visual navigation and
Atari games.  \citet{millidge2020deep} connects active inference
with variational policy gradients, and
\citet{lanillos2021active} surveys active inference in robotics.
\citet{delavari2025perceptual} demonstrate active inference for
lateral vehicle control via perceptual motor learning.  These
works focus on single behaviors; we additionally show that runtime
weight modulation of EFE components can produce multiple behaviors
from a shared world model.

\textbf{Epistemic uncertainty in planning.}
Deep ensembles~\citep{lakshminarayanan2017simple} provide
calibrated uncertainty via disagreement among independently trained
networks and are widely used in model-based
RL~\citep{chua2018deep}. Evidential deep
learning~\citep{sensoy2018evidential, amini2020deep} places
distributional priors over network outputs to separate aleatoric
from epistemic uncertainty in a single forward pass.
\citet{sekar2020planning} plan to explore via world model
uncertainty. These methods quantify epistemic uncertainty at
\emph{observation time}, but assume that the resulting signals
transfer faithfully into model-based planning rollouts. Our
results suggest that this assumption may not hold for RSSM-based
active inference: posterior epistemic competence does not appear
to guarantee prior (imagination) competence, even when the
representation encodes the task-relevant signal with high fidelity.

\textbf{Precision in active inference.}
Precision-weighting---modulating the relative influence of
prediction errors---is central to biological active
inference~\citep{parr2022active}. Computational implementations
typically use fixed weights; we demonstrate runtime modulation of
EFE component weights to produce qualitatively different behaviors
from a shared world model.

%----------------------------------------------------------------------
\section{Background}

\subsection{Active Inference and EFE}

Under active inference, an agent selects policies $\pi$ that
minimize Expected Free Energy:
% \begin{equation}
%     G(\pi) = \sum_t \gamma^t \Big[
%         \underbrace{\mathbb{E}_q[-\log p_{\text{pref}}(z_t)]}_{\text{instrumental}}
%         + \underbrace{H[p(o_t | s_t)]}_{\text{epistemic}}
%     \Big]
% \end{equation}
% where the instrumental term drives the agent toward preferred
% states $p_{\text{pref}}$ and the epistemic term promotes
% information-seeking. We define $p_{\text{pref}}$ in latent space
% rather than observation space because the RSSM planning rollout
% operates entirely in latent space, making latent-space preferences
% directly compatible with the planner without an additional decoding
% step. Precision-weighting modulates the relative influence of
% these terms based on their estimated reliability.
% \paragraph{A note on ``epistemic.''}
% We use this term in two distinct senses throughout the paper:
% (i)~the \emph{epistemic value} $G_e$ in the EFE, which quantifies
% expected information gain from actions (a standard AIF concept),
% and (ii)~the \emph{posterior-prior epistemic gap}, which
% quantifies how much task-relevant signal the transition model
% loses when rolling out imagined states without observations. The
% first is a planning objective; the second is a diagnostic of the
% world model's imagination fidelity.
\begin{equation}
    G(\pi) = \sum_t \gamma^t \Big[
        \underbrace{\mathbb{E}_q[-\log p_{\text{pref}}(z_t)]}_{\text{instrumental (risk)}}
        + \underbrace{H[p(o_t | s_t)]}_{\text{ambiguity}}
    \Big],
\end{equation}
where the instrumental term drives the agent toward preferred
states $p_{\text{pref}}$ and the ambiguity term penalizes states
where observations are unpredictable from the latent state.
The full epistemic value---mutual information
$I(o;s) = H[q(o)] - H[p(o|s)]$---further includes an
expected-novelty term, but neither decomposition is computed
directly in our implementation; instead, Equation~\eqref{eq:efe}
uses ensemble disagreement as an empirical proxy for epistemic
uncertainty.
We define $p_{\text{pref}}$ in latent space rather than
observation space because the RSSM planning rollout operates
entirely in latent space, making latent-space preferences
directly compatible with the planner without an additional
decoding step. Precision-weighting modulates the relative
influence of these terms based on their estimated reliability.
\paragraph{A note on ``epistemic.''}
We use this term in two distinct senses throughout the paper:
(i)~the \emph{epistemic value} $G_e$ in our EFE as seen in~\eqref{eq:efe}, which uses ensemble disagreement as an
empirical proxy for information gain, and
(ii)~the \emph{posterior-prior epistemic gap}, which quantifies
how much task-relevant signal the transition model loses when
rolling out imagined states without observations. The first is a
planning-time uncertainty signal; the second is a diagnostic of
the world model's imagination fidelity.

\subsection{RSSM World Model}

The RSSM maintains deterministic state $h_t \in \mathbb{R}^{256}$
(GRU) and stochastic state $z_t \in \mathbb{R}^{64}$ (Gaussian):
\begin{align}
    \text{Posterior:} \quad z_t &\sim q(z_t | h_t, o_t) & \text{(filtering)} \\
    \text{Prior:} \quad \hat{z}_t &\sim p(z_t | h_t) & \text{(imagination)}
\end{align}
The concatenated feature $[h_t; z_t] \in \mathbb{R}^{320}$ serves
as the latent state for decoding and planning. During planning,
only the prior is available---this distinction becomes critical in
Section~\ref{sec:vision}.

%----------------------------------------------------------------------
\section{Method}

\subsection{Architecture}

The world model comprises a ConvEncoder (4-layer CNN processing
64$\times$64 RGB + 4D proprioceptive state
$[\text{speed}, \text{steer}, h_{\text{err}}, c_{\text{err}}]$
$\rightarrow$ 256D embedding), RSSM (deter=256, stoch=64),
ObsDecoder (image reconstruction), StateDecoder (4D state
prediction), and 5-head EnsembleTransitionHeads (epistemic
uncertainty). Training minimizes Variational Free Energy (VFE)
with dual KL divergence~\citep{hafner2023mastering}, image and
state reconstruction losses, on 96K frames of expert driving from
CARLA Towns 04 and 06.

\subsection{Multi-Component EFE}

Our EFE scoring evaluates imagined trajectories from iCEM (500
samples, 50 elites, 5 iterations, horizon 12--15) with four
components:
\begin{equation}
    G(\tau) = \sum_{t=1}^{H} \gamma^t \Big[
    \beta_i G_i(t) - \beta_e G_e(t) + \beta_s G_s(t) + \beta_o G_o(t)
    \Big]
\label{eq:efe}
\end{equation}

\textbf{Instrumental value} $G_i$: Cross-entropy
$\mathbb{E}_q[-\log p_{\text{pref}}(z_t)]$ between imagined
latents and a Gaussian Mixture Model (GMM) preference fitted on expert driving latents
($K{=}5$, 64D). Z-scored across the CEM population per timestep.

\textbf{Epistemic value} $G_e$: Ensemble disagreement among 5
transition heads. Because $G$ is minimized, the negative sign on
$\beta_e G_e$ in \eqref{eq:efe} maximizes epistemic value,
promoting information-seeking trajectories.

\textbf{State-space value} $G_s$: Decoded heading error and
crosstrack error from the state decoder, penalizing deviation from
lane center: $G_s(t) = \min(\hat{h}_t^2 + \hat{c}_t^2, 4.0)$.
Not z-scored---state decoder predictions during imagination are
noisy, and z-scoring amplifies this noise.

\textbf{Obstacle penalty} $G_o$: Runtime lane-change incentive
using detected obstacle proximity and decoded crosstrack shift.
Penalizes imagined trajectories that maintain the current lateral
position when an obstacle is ahead.

\subsection{Runtime Weight Modulation}
\label{sec:precision}

The $\beta$ weights are modulated at runtime based on prediction
errors and context.  These hand-designed schedules approximate
precision-weighting~\citep{parr2022active} without formal
optimality guarantees.  Three rules apply: (1)~\emph{state error
modulation}---when $e = h_{\text{err}}^2 + c_{\text{err}}^2 > 0.3$,
$\beta_s$ increases (up to $4\times$) and $\beta_i$ decreases
proportionally; (2)~\emph{obstacle proximity modulation}---when
proximity $\rho > 0.1$, $\beta_i$ is suppressed
($\beta_i \leftarrow \beta_i \cdot \max(0.2, 1{-}0.8\rho)$) to
let $G_o$ dominate; (3)~\emph{post-evasion centering}---after
obstacle clearance, $\beta_s$ is temporarily set to 0.5, pulling
the ego back toward lane center.

\begin{table}[t]
\centering\small
\caption{Task-conditioned EFE component weights. Same world model
checkpoint; only $\beta$ weights and horizon differ.}
\label{tab:config}
\begin{tabular}{lcc}
\toprule
\textbf{Parameter} & \textbf{Task A} & \textbf{Task B} \\
\midrule
$\beta_{\text{instrumental}}$ & 1.0 & 1.0 \\
$\beta_{\text{epistemic}}$ & 0.1 & 0.1 \\
$\beta_{\text{state}}$ & 0.5 & 0.0 \\
$\beta_{\text{obstacle}}$ & 0.0 & 40.0 \\
Horizon $H$ & 12 & 15 \\
\bottomrule
\end{tabular}
\vspace{-2mm}
\end{table}

Table~\ref{tab:config} shows the key difference: Task~A uses
$\beta_s{=}0.5$ for continuous centering while Task~B sets
$\beta_s{=}0.0$ to allow lateral deviation during lane changes,
with $\beta_o{=}40.0$ for obstacle avoidance.

%----------------------------------------------------------------------
\section{Behavioral Results (Privileged System Context)}

All experiments use CARLA 0.9.16 with synchronous mode at 20~FPS.
Mean Lateral Deviation (MLD) measures lane-center accuracy;
Success Rate (SR) counts collision-free episodes.

\subsection{Task A: Lane Keeping}

On Town04 moderate curves (378m, 0$^\circ$--85$^\circ${} heading
change, 3 episodes), the agent achieves $81.2 \pm 1.8$\% average
route completion with $0.63 \pm 0.13$\,m MLD.
The state-space penalty ($\beta_s{=}0.5$) provides a
Gaussian preference $\tilde{p}(s) = \mathcal{N}(0, I)$ over
heading and crosstrack errors. The agent terminates at a sharp
curve beyond the training distribution ($>$85$^\circ$).

\subsection{Task B: Obstacle Avoidance}

On Town06 highway routes (373--518m, 2--3 static obstacles per
route, using privileged obstacle coordinates), the agent avoids
28/28 obstacles across 11 episodes on three route configurations
(Table~\ref{tab:results}). We note the limited scale: 11 episodes
on 3 fixed routes with static obstacles.

\begin{table}[t]
\centering\small
\caption{Obstacle avoidance results using \emph{privileged}
runtime coordinates (mean $\pm$ std over episodes).
SR: Success Rate; LC: Lane Changes per episode;
MLD: Mean Lateral Deviation (m).}
\label{tab:results}
\setlength{\tabcolsep}{4pt}
\begin{tabular}{lccccc}
\toprule
\textbf{Route} & \textbf{Eps} & \textbf{SR} & \textbf{Avoid} & \textbf{MLD} & \textbf{LC} \\
\midrule
R0 (373m, 2\,obs) & 5 & 100\% & 10/10 & $0.67{\scriptstyle\pm.05}$ & 6.0 \\
R1 (518m, 3\,obs) & 3 & 100\% & 9/9 & $0.66{\scriptstyle\pm.03}$ & 6.0 \\
R2 (518m, 3\,alt) & 3 & 100\% & 9/9 & $0.67{\scriptstyle\pm.02}$ & 6.3 \\
\midrule
\textbf{Total} & 11 & \textbf{100\%} & \textbf{28/28} & $0.66{\scriptstyle\pm.04}$ & 6.1 \\
\bottomrule
\end{tabular}
\end{table}

Each obstacle is avoided via an active lane-change maneuver
($\sim$6 lane changes per episode).
Post-evasion centering ($\beta_s \rightarrow 0.5$ after obstacle
clearance) returns the ego toward the route lane between
obstacles.  Both tasks use the same trained world model
checkpoint; the behavioral difference arises entirely from
$\beta$ configuration (Table~\ref{tab:config}).

\subsection{Ablation}

Table~\ref{tab:ablation} shows Route~1 ablation results.

\begin{table}[t]
\centering\small
\caption{Ablation on R1 (518m, 3 obstacles, 3 episodes each).
Same 64$\times$64 model throughout.  ``w/o privileged'' removes
all six privileged mechanisms simultaneously (not individually
isolated); Compl.: route completion.}
\label{tab:ablation}
\begin{tabular}{lcccc}
\toprule
\textbf{Variant} & \textbf{SR} & \textbf{Avoid} & \textbf{MLD} & \textbf{Compl.} \\
\midrule
Full system & 100\% & 9/9 & 0.66 & 99.2\% \\
w/o privileged & 0\% & 0/9 & 0.28 & 43.8\% \\
$+$ Centering & 100\% & 9/9 & 0.52 & 99.1\% \\
\bottomrule
\end{tabular}
\vspace{1mm}

\small{``w/o privileged'': all 6 obstacle mechanisms removed
(obstacle penalty $G_o$, proximity computation, evasion direction,
reflexive steer, directional CEM penalty, motor primitives).
``$+$ Centering'': full system with post-evasion $\beta_s{=}0.5$.}
\end{table}

Removing all privileged mechanisms yields 0\% avoidance despite
reasonable driving (0.28\,m MLD), confirming their necessity.
Post-evasion centering improves MLD (0.66$\rightarrow$0.52)
without degrading avoidance.

%----------------------------------------------------------------------
\section{Toward Vision-Based Obstacle Detection}
\label{sec:vision}

The obstacle avoidance results above rely on runtime obstacle
coordinates. We investigate whether this information can emerge
from the latent space, removing all six privileged mechanisms
(proximity computation, evasion direction, reflexive steer
override, motor primitives, directional CEM penalty, obstacle
EFE penalty).

\subsection{Preference Gap Analysis}

We fit a GMM on obstacle-free expert latents and measure the gap
$\Delta = \log p(z_{\text{clean}}) - \log p(z_{\text{obs}})$.
A positive gap would enable EFE-driven avoidance.

\begin{table}[h]
\centering\small
\caption{Preference gap $\Delta$ across modalities and methods.
Top rows vary input modality (separate world models); bottom rows
vary the preference objective (same 64$\times$64 model).
$\alpha$: contrastive down-weighting factor.}
\label{tab:gap}
\begin{tabular}{lccc}
\toprule
\textbf{Method} & \textbf{$\Delta$} & \textbf{MLD} & \textbf{Avoid} \\
\midrule
\multicolumn{4}{l}{\emph{Single GMM (different modalities)}} \\
\quad 64$\times$64 RGB & $-$24.6 & 0.31 & 0\% \\
\quad 128$\times$128 RGB & $-$2.3 & 0.17 & 0\% \\
\quad 64$\times$64 RGBD & $-$68.1 & --- & 0\% \\
\midrule
\multicolumn{4}{l}{\emph{Contrastive preference ($\alpha{=}0.2$)}} \\
\quad Without aux head & +1.1 & 0.17 & 0\% \\
\quad With aux head & +133.1 & 0.61 & 0\% \\
\bottomrule
\end{tabular}
\vspace{1mm}

\small{$\Delta$: effective nats (positive = correct). Aux head:
binary obstacle prediction during training only.}
\end{table}

\subsection{Contrastive Preference}

We replace the single GMM with a log-ratio model:
$\log p_{\text{pref}}(z) = \log p_{\text{clean}}(z)
- \alpha \log p_{\text{avoid}}(z)$, where $\alpha$ controls
how strongly obstacle-present latents are down-weighted.
This achieves correct posterior separation (37 nats raw at
$\alpha{=}1.0$, 1.1 effective nats at $\alpha{=}0.2$), but
avoidance remains 0\%.

\subsection{Auxiliary Prediction Head}

To determine whether the bottleneck is the representation or the
transition model, we add a supervised obstacle prediction head
(MLP: 320$\rightarrow$128$\rightarrow$1, BCE loss,
$\beta_{\text{aux}}{=}1.0$) during training only. On the training
distribution, this achieves near-perfect posterior separation
(0.000 vs 1.000 on in-distribution task labels) and amplifies the
contrastive gap to 276 nats (133.1 effective). Cross-dataset
probing (clean vs.\ obstacle-present data, Table~\ref{tab:probe})
yields AUC 0.85---strong but imperfect, reflecting distribution
shift between training and probing data. Yet avoidance remains at
\textbf{0\%}.

\subsection{The Posterior-Prior Epistemic Gap}

\textbf{Protocol.}  We collect 100 obstacle-free and 100
obstacle-present sequences from the evaluation routes, encode each
with the RSSM posterior (using ground-truth observations), and
record the auxiliary-head logit at the obstacle encounter frame.
For imagined priors, we seed the RSSM at the same posterior state
and unroll via \texttt{img\_step} for $k{=}1$ to 15 steps using
the agent's own planned actions, applying the same frozen
auxiliary head at each depth (Table~\ref{tab:probe}).

\begin{figure}[t]
\centering
\includegraphics[width=\columnwidth]{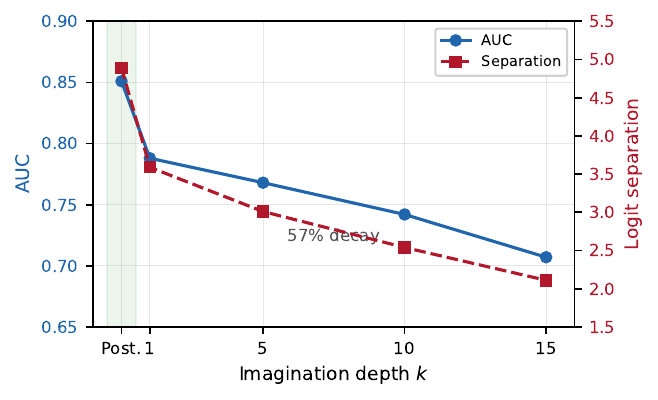}
\caption{Posterior-prior epistemic gap across imagination depth.
AUC (left axis) and logit separation (right axis) both decay
monotonically as the RSSM rolls out imagined states without
observations. AUC drops from 0.85 (posterior) to 0.71 ($k{=}15$),
with logit separation decaying 57\% over 15 steps. The sharpest
drop occurs at $k{=}1$.}
\label{fig:probe_decay}
\end{figure}

\textbf{Caveat.}  The auxiliary head is trained on posterior
states; evaluating it on prior states introduces a
distribution-shift confound.  Two observations mitigate this
concern: (1)~a fixed posterior--prior distribution offset would
produce a \emph{constant} AUC drop at $k{=}1$ with no further
depth dependence, whereas we observe monotonic decay from
$k{=}1$ through $k{=}15$
(Figure~\ref{fig:probe_decay}); (2)~the sharpest drop occurs
at $k{=}1$ (separation 4.89$\rightarrow$3.59), consistent with
the transition model's single-step information loss compounding
over depth.  Nevertheless, a probe retrained on imagined prior states is
the clean next test; our setup lacks ground-truth obstacle
labels aligned with imagination, which we leave to future work.

\begin{table}[t]
\centering\small
\caption{Obstacle probe on posterior vs.\ imagined prior states.
AUC and logit separation (mean obstacle $-$ mean clean) decay with
imagination depth.}
\label{tab:probe}
\begin{tabular}{lcc}
\toprule
\textbf{State} & \textbf{AUC} & \textbf{Separation} \\
\midrule
Posterior & 0.851 & 4.89 \\
Prior $k{=}1$ & 0.788 & 3.59 \\
Prior $k{=}5$ & 0.768 & 3.01 \\
Prior $k{=}10$ & 0.742 & 2.54 \\
Prior $k{=}15$ & 0.707 & 2.11 \\
\bottomrule
\end{tabular}
\vspace{1mm}

\small{200 sequences (balanced), aux-head model. Separation
decays 57\% over 15 imagination steps.}
\end{table}

The posterior encodes obstacle presence with AUC 0.85, but
imagination progressively erodes this signal: AUC drops to
0.71 and logit separation decays 57\% over 15 steps
(Figure~\ref{fig:probe_decay}).  This is consistent with the
RSSM transition model $f(h_t, z_t, a_t) \rightarrow h_{t+1}$
failing to propagate obstacle information through
action-conditioned rollouts (subject to the distribution-shift
caveat above).  The sharpest drop occurs at $k{=}1$ (separation
4.89$\rightarrow$3.59), indicating that even a single
imagination step substantially degrades obstacle encoding.

\textbf{Why $G_e$ does not help.}
The ensemble-based epistemic term $G_e$ measures disagreement
among five transition heads.  During the same imagined rollouts,
we measure whether $G_e$ can distinguish obstacle-present from
obstacle-free trajectories: the AUC of ensemble disagreement
as a binary obstacle classifier is 0.52 at $k{=}1$ and 0.51 at
$k{=}15$ (chance level), with Cohen's $d < 0.1$ at all depths.
The ensemble heads agree on \emph{how} obstacles are
forgotten---they share the same blind spot.  Consequently, $G_e$
provides no planning-time signal that obstacle information has
been lost, making the failure an unknown unknown.

Combined with the 0\% avoidance in Table~\ref{tab:gap}, this
shows that the residual prior signal (AUC 0.71) is insufficient
to drive EFE-based obstacle avoidance through CEM planning.

%----------------------------------------------------------------------
\section{Discussion}

\textbf{Task-conditioned weighting over a shared world model.}
Different $\beta$ configurations over the same world model produce
lane keeping, obstacle avoidance (with privileged coordinates),
and post-evasion centering---consistent with precision-weighting
in active inference~\citep{parr2022active}, though our schedules
are hand-designed.

\textbf{Posterior $\neq$ prior epistemic competence.}
Direct probing (Table~\ref{tab:probe},
Figure~\ref{fig:probe_decay}) reveals a measurable dissociation
in this single-seed study: the posterior encodes obstacles at
AUC 0.85, but imagination erodes this to AUC 0.71 over 15 steps
(57\% logit-separation loss).  This suggests that planning quality
may be bounded by the transition model's ability to propagate
task-relevant information through imagination, not solely by the
representation's expressiveness.  We note the distribution-shift
caveat (Section~\ref{sec:vision}): because the probe is trained
on posterior states, some of the measured decay may reflect
probe--distribution mismatch rather than pure signal loss.
Whether this limitation extends to other RSSM-based systems (e.g.,
DreamerV3~\citep{hafner2023mastering}) is an open question.

\textbf{An unflagged imagination failure.}
In this setup, the gap constitutes an operationally ``unknown
unknown'': ensemble disagreement ($G_e$) evaluated as a binary
obstacle classifier yields AUC 0.52 at $k{=}1$ and 0.51 at
$k{=}15$ (chance level; Cohen's $d < 0.1$), confirming that the
planner's own uncertainty signal does not distinguish safe from
dangerous imagined futures.  From the planner's perspective,
high-collision and safe trajectories are
indistinguishable---a failure mode that cannot be detected without
external observation.  This aligns with the EIML workshop theme:
in this case study, the agent does not know what it does not know
during planning.

\textbf{Limitations.}
This is a single-seed case study: one trained world model, 11
episodes across 3 fixed highway routes with static obstacles, and
no training-seed variation.  Results should be interpreted as a
diagnostic case study, not as broadly generalizable claims.
The auxiliary-head probe is trained on posterior states and
evaluated on prior states, introducing a distribution-shift
confound (Section~\ref{sec:vision}); the monotonic
depth-dependent decay is suggestive but not conclusive without a
matched prior-state probe.  The ``w/o privileged'' ablation
(Table~\ref{tab:ablation}) removes all six privileged mechanisms
simultaneously, so it confirms necessity of the bundle but does
not isolate individual components.
Obstacle avoidance relies on runtime obstacle coordinates---a form
of privileged information---rather than learned perception.
The component weighting schedules are hand-designed, not learned.
We do not compare against non-AIF planners (e.g., MPC with
hand-designed cost) or imitation baselines.  Sharp curves beyond
$\sim$85$^\circ${} exceed the training distribution, and no
junctions or dynamic traffic are tested.

\textbf{Future directions.}
Closing the posterior-prior gap requires transition models that
learn object-conditioned dynamics. Attention-based transitions
that selectively propagate object features through imagination,
or object-centric world models
(SLATE~\citep{singh2022illiterate},
SAVi~\citep{kipf2022conditional}) that decompose scenes into
slots, may enable the transition model to predict how obstacles
move through the visual field as a function of steering actions.

%----------------------------------------------------------------------
\section{Conclusion}

In a single-seed case study, we measured a posterior-prior gap
in RSSM-based planning: a supervised obstacle probe drops from
AUC 0.85 (posterior) to 0.71 (15-step imagination), with 57\%
logit-separation decay.  Avoidance remains at 0\% without
privileged coordinates, and the planner's ensemble-based
epistemic term does not flag the loss---making the failure an
unknown unknown at planning time.  While the probe's
distribution-shift caveat prevents a fully conclusive attribution,
the monotonic depth-dependent decay points to the transition model
as the primary bottleneck.  Closing this gap---through
object-conditioned transitions, attention-based imagination, or
imagination-aware uncertainty estimation---is a key challenge for
world-model-based agents requiring object-level reasoning.

As a secondary contribution, task-conditioned weighting of EFE
components over the same RSSM produced lane keeping
($81.2 \pm 1.8$\% route completion) and obstacle avoidance
(28/28 with privileged coordinates) from a single objective.

%----------------------------------------------------------------------
\bibliographystyle{plainnat}
\bibliography{references}

@article{friston2010free,
  title={The free-energy principle: a unified brain theory?},
  author={Friston, Karl},
  journal={Nature Reviews Neuroscience},
  volume={11},
  number={2},
  pages={127--138},
  year={2010}
}

@book{parr2022active,
  title={Active Inference: The Free Energy Principle in Mind, Brain, and Behavior},
  author={Parr, Thomas and Pezzulo, Giovanni and Friston, Karl J},
  year={2022},
  publisher={MIT Press}
}

@article{hafner2023mastering,
  title={Mastering diverse domains through world models},
  author={Hafner, Danijar and Pasukonis, Jurgis and Ba, Jimmy and Lillicrap, Timothy},
  journal={arXiv preprint arXiv:2301.04104},
  year={2023}
}

@inproceedings{dosovitskiy2017carla,
  title={{CARLA}: An open urban driving simulator},
  author={Dosovitskiy, Alexey and Ros, German and Codevilla, Felipe and Lopez, Antonio and Koltun, Vladlen},
  booktitle={Conference on Robot Learning},
  pages={1--16},
  year={2017}
}

@article{catal2020learning,
  title={Learning perception and planning with deep active inference},
  author={Catal, Ozan and Wauthier, Samuel and De Boom, Cedric and Verbelen, Tim and Dhoedt, Bart},
  journal={arXiv preprint arXiv:2001.11841},
  year={2020}
}

@inproceedings{fountas2020deep,
  title={Deep active inference agents using {Monte-Carlo} methods},
  author={Fountas, Zafeirios and Sajid, Noor and Mediano, Pedro AM and Friston, Karl},
  booktitle={Advances in Neural Information Processing Systems},
  volume={33},
  pages={11662--11675},
  year={2020}
}

@article{millidge2020deep,
  title={Deep active inference as variational policy gradients},
  author={Millidge, Beren},
  journal={Journal of Mathematical Psychology},
  volume={96},
  pages={102348},
  year={2020}
}

@article{delavari2025perceptual,
  title={Perceptual Motor Learning with Active Inference Framework for Robust Lateral Control},
  author={Delavari, Elahe and Moore, John and Hong, Junho and Kwon, Jaerock},
  journal={arXiv preprint arXiv:2503.01676},
  year={2025}
}

@article{lanillos2021active,
  title={Active inference in robotics and artificial agents: Survey and challenges},
  author={Lanillos, Pablo and others},
  journal={arXiv preprint arXiv:2112.01871},
  year={2021}
}

@inproceedings{hu2022model,
  title={Model-based imitation learning for urban driving},
  author={Hu, Anthony and others},
  booktitle={Advances in Neural Information Processing Systems},
  volume={35},
  year={2022}
}

@inproceedings{chua2018deep,
  title={Deep reinforcement learning in a handful of trials using probabilistic dynamics models},
  author={Chua, Kurtland and Calandra, Roberto and McAllister, Rowan and Levine, Sergey},
  booktitle={Advances in Neural Information Processing Systems},
  volume={31},
  year={2018}
}

@inproceedings{sekar2020planning,
  title={Planning to explore via self-supervised world models},
  author={Sekar, Ramanan and Rybkin, Oleh and Daniilidis, Kostas and Abbeel, Pieter and Hafner, Danijar and Pathak, Deepak},
  booktitle={International Conference on Machine Learning},
  pages={8583--8592},
  year={2020}
}

@inproceedings{singh2022illiterate,
  title={Illiterate {DALL-E} learns to compose},
  author={Singh, Gautam and Deng, Fei and Ahn, Sungjin},
  booktitle={International Conference on Learning Representations},
  year={2022}
}

@inproceedings{pinneri2021sample,
  title={Sample-efficient cross-entropy method for real-time planning},
  author={Pinneri, Cristina and Sawant, Shambhuraj and Blaes, Sebastian and Achterhold, Jan and Stueckler, Joerg and Rolinek, Michal and Martius, Georg},
  booktitle={Conference on Robot Learning},
  pages={1049--1065},
  year={2021}
}

@inproceedings{kipf2022conditional,
  title={Conditional object-centric learning from video},
  author={Kipf, Thomas and Elsayed, Gamaleldin F and Mahendran, Aravindh and Stone, Austin and Sabour, Sara and Heigold, Georg and Jonschkowski, Rico and Dosovitskiy, Alexey and Greff, Klaus},
  booktitle={International Conference on Learning Representations},
  year={2022}
}

@inproceedings{lakshminarayanan2017simple,
  title={Simple and scalable predictive uncertainty estimation using deep ensembles},
  author={Lakshminarayanan, Balaji and Pritzel, Alexander and Blundell, Charles},
  booktitle={Advances in Neural Information Processing Systems},
  volume={30},
  year={2017}
}

@inproceedings{amini2020deep,
  title={Deep evidential regression},
  author={Amini, Alexander and Schwarting, Wilko and Soleimany, Amin and Rus, Daniela},
  booktitle={Advances in Neural Information Processing Systems},
  volume={33},
  pages={14927--14937},
  year={2020}
}

@inproceedings{sensoy2018evidential,
  title={Evidential deep learning to quantify classification uncertainty},
  author={Sensoy, Murat and Kaplan, Lance and Kandemir, Melih},
  booktitle={Advances in Neural Information Processing Systems},
  volume={31},
  year={2018}
}

@article{chen2022transdreamer,
  title={Trans{D}reamer: Reinforcement Learning with Transformer World Models},
  author={Chen, Chang and Wu, Yi-Fu and Yoon, Jaesik and Ahn, Sungjin},
  journal={arXiv preprint arXiv:2202.09481},
  year={2022}
}

@inproceedings{micheli2023transformers,
  title={Transformers are Sample-Efficient World Models},
  author={Micheli, Vincent and Alonso, Eloi and Fleuret, Fran{\c{c}}ois},
  booktitle={The Eleventh International Conference on Learning Representations},
  year={2023},
  url={https://openreview.net/forum?id=vhFu1Acb0xb}
}

\end{document}